\documentclass[11pt,a4paper]{article}
\usepackage[hyperref]{emnlp2020}
\usepackage{times}
\usepackage{latexsym}
\usepackage{subfigure}
\usepackage{multirow}
\usepackage{graphicx}
\usepackage{amsfonts}

\usepackage{enumitem}
\usepackage{microtype}

\aclfinalcopy % Uncomment this line for the final submission
\newcommand{\tabincell}[2]{\begin{tabular}{@{}#1@{}}#2\end{tabular}}

\title{Pretrained Language Models for Dialogue Generation\\ with Multiple Input Sources }

\author{Yu Cao \textsuperscript{1}\thanks{~~This work was done during Yu Cao's internship in Tencent AI LAB, Shenzhen.} ,~~~~Wei Bi \textsuperscript{2},~~~~Meng Fang \textsuperscript{3},~~~~Dacheng Tao \textsuperscript{1}\\
$^1$ UBTECH Sydney AI Center, School of Computer Science, Faculty of Engineering, \\
The University of Sydney, Australia\\
$^2$ Tencent AI LAB, China,~~~~$^3$ Tencent Robotics X, China\\
\tt ycao8647@uni.sydney.edu.au, victoriabi@tencent.com, \\ \tt mfang@tencent.com, dacheng.tao@sydney.edu.au}
\date{}

\begin{document}
\maketitle
\begin{abstract}
Large-scale pretrained language models have achieved outstanding performance on natural language understanding tasks. However, it is still under investigating how to apply them to dialogue generation tasks, especially those with responses conditioned on multiple sources. Previous work simply concatenates all input sources or averages information from different input sources. 
In this work, we study dialogue models with multiple input sources adapted from the pretrained language model GPT2.
We explore various methods to fuse multiple separate attention information corresponding to different sources.
Our experimental results show that proper fusion methods deliver higher relevance with dialogue history than simple fusion baselines.
% \footnote{***rewrite the expt findings and wrap up the ref.}
%the promotion brought by these refined approaches, and analysis also explains why they take effect under some conditions.
% Personalized dialogue generation requires correlation to both history and specific profile, so how model conditions on the extra information is important. Previous methods handle it by concatenating them as a single input or averaging encoded states from different sources. In this paper, based on a encoder-decoder architecture with pre-trained GPT2 model as encoder and decoder, we utilizes various kinds of operations to fuse attention information from different sources, including static ones and dynamic one with trainable parameters. 
\end{abstract}

\section{Introduction}

Large-scale pretrained language models~\cite{bert,gpt,gpt2} have achieved outstanding performance on various natural language understanding tasks~\cite{young2018recent,bert_nlu}.
%make transfer learning from a large corpus to a specific NLP task possible. 
Researchers have then utilized them in dialogue generation tasks~\cite{gpt2_generation, gpt_generation, dialogpt}. Many of them simply concatenate the input dialogue history and the output response in finetuning, since the pretrained language model only accepts a single sequence as input.
However, dialogue generation tasks may involve multiple input sources simultaneously. For example, in personalized or knowledge-grounded dialogue generation~\cite{twitterpersona,personachat,wow}, a response is generated conditioned on both dialogue history and an auxiliary user profile or knowledge article.
Despite simple concatenation of all input sources, an important question arises on how we can better adapt a single-input pretrained language model to a multi-input dialogue generation task.

%In previous works, TransferTransfo~\cite{transfertransfo} applies OpenAI GPT on personalized dialog generation by using the concatenation of personality, dialog history and current response label as input. Although it outperforms end-to-end models, such simple strategy may not pose enough importance on extra information.
%To address this problem, 
Some previous work forms an encoder-decoder architecture 
with both encoder and decoder duplicated from a pretrained language model~\cite{gpt-persona,persona_sparse}.
Recently, BART~\cite{bart} even obtain a complete pretrained model under this architecture directly.
Taking personalized dialogue generation~\cite{personachat} as an example, we can treat persona information, dialogue history and previous generated tokens as three different input sources.  
The former two will be encoded firstly and then combined with the last one in the decoder.
In \citeauthor{gpt-persona}~\citeyear{gpt-persona},
the multi-head attention layer in the decoder is copied three
times for each input source and
%Its mean pooling on multiple attentions cannot ensure it is optimal under various context conditions despite better than single input. 
mean pooling is used to average results from multiple attentions. 
This encoder-decoder adaptation is shown to outperform simple concatenation.

However, when dialogue history gets longer, this model tends to use less information of each dialogue history utterance to predict the next token.
\citeauthor{persona_sparse}~\citeyear{persona_sparse} add an extra weight predictor to combine multiple attention information, but they do not perform experiments using publicly released pretrained models, nor on public datasets, making their results not directly comparable to other work.

In this work, we build our dialogue model on the encoder-decoder architecture adapted from the pretrained language model GPT2~\cite{gpt2}. Our main contribution is to empirically study the attention fusion methods for multiple information sources in each decoder layer. Three kinds of methods are explored in total.
Our experimental results show performance improvements on both automatic and human evaluations by using proper attention fusion methods, compared to baselines using concatenation or mean pooling.
% And with the more powerful GPT-2 model and multi-task learning loss, even our raw model also outperforms previous baselines.

%Our contribution can be concluded as:
% \noindent 1) We propose NLG models with a series of refined fusion approaches to better combine attention from different input sources, based on a multi-input encoder-decoder GPT2 model.

% \noindent 2) Our experiments show the improvement compared to previous models. Related analysis and case study also illustrates why new attention fusion methods works.
%\begin{itemize}[wide=0\parindent, noitemsep, topsep=0]
%    \item We propose NLG models with a series of refined fusion approaches to better combine attention from different input sources, based on a multi-input encoder-decoder GPT2 model.
%    \item Our experiments show the improvement compared to previous models. Related analysis and case study also illustrates why these attention fusion methods work better.
%\end{itemize}

\section{Model}
\subsection{The Encoder-Decoder Architecture}
\begin{figure}[!t]
\setlength{\abovecaptionskip}{-0cm}
\setlength{\belowcaptionskip}{-0.1cm}
    \centering
    \subfigure[The encoder-decoder architecture.]{
        \includegraphics[width=0.48\textwidth]{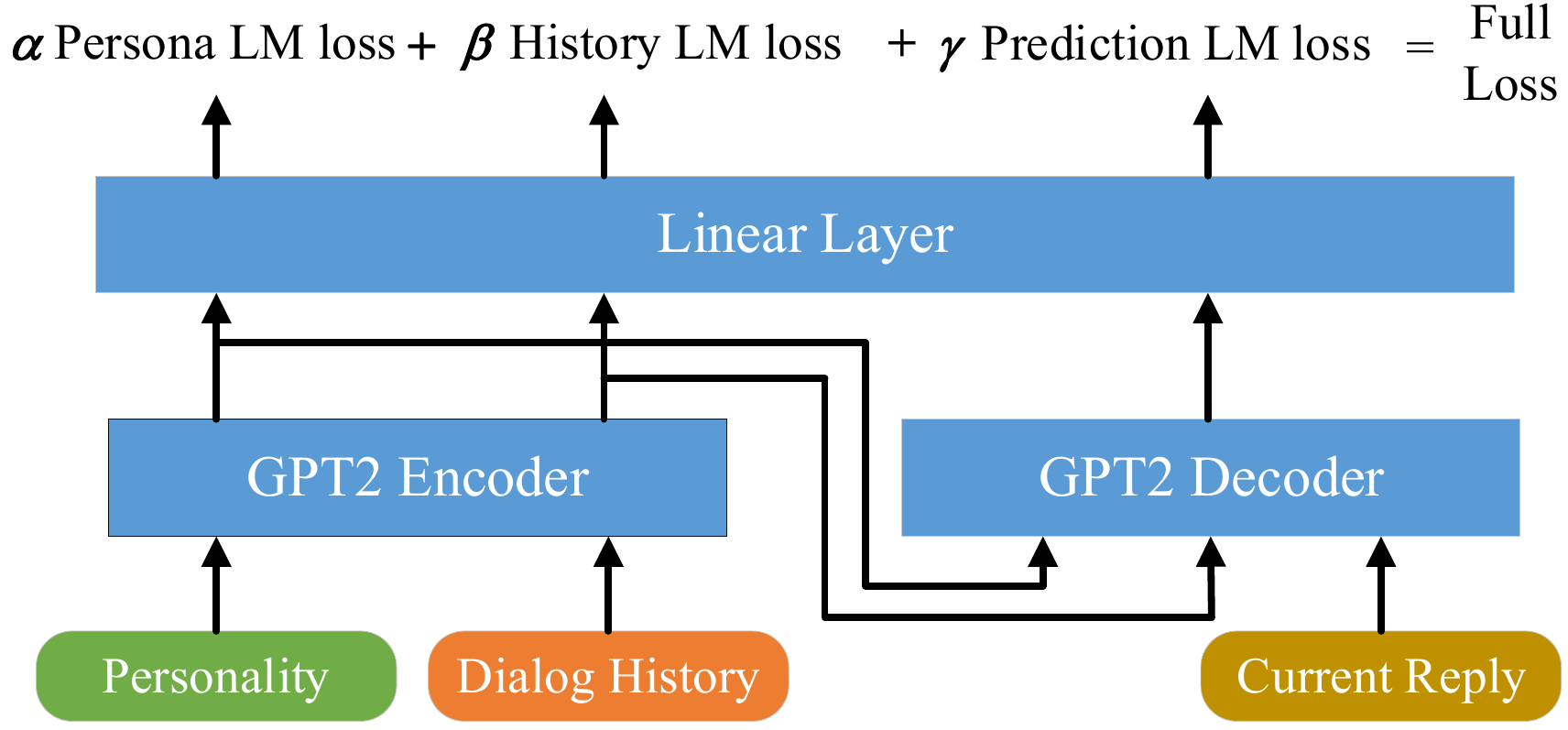}
        \label{fig:framework}
    }
    \subfigure[Details of each transformer block in decoder.]{
        \includegraphics[width=0.48\textwidth]{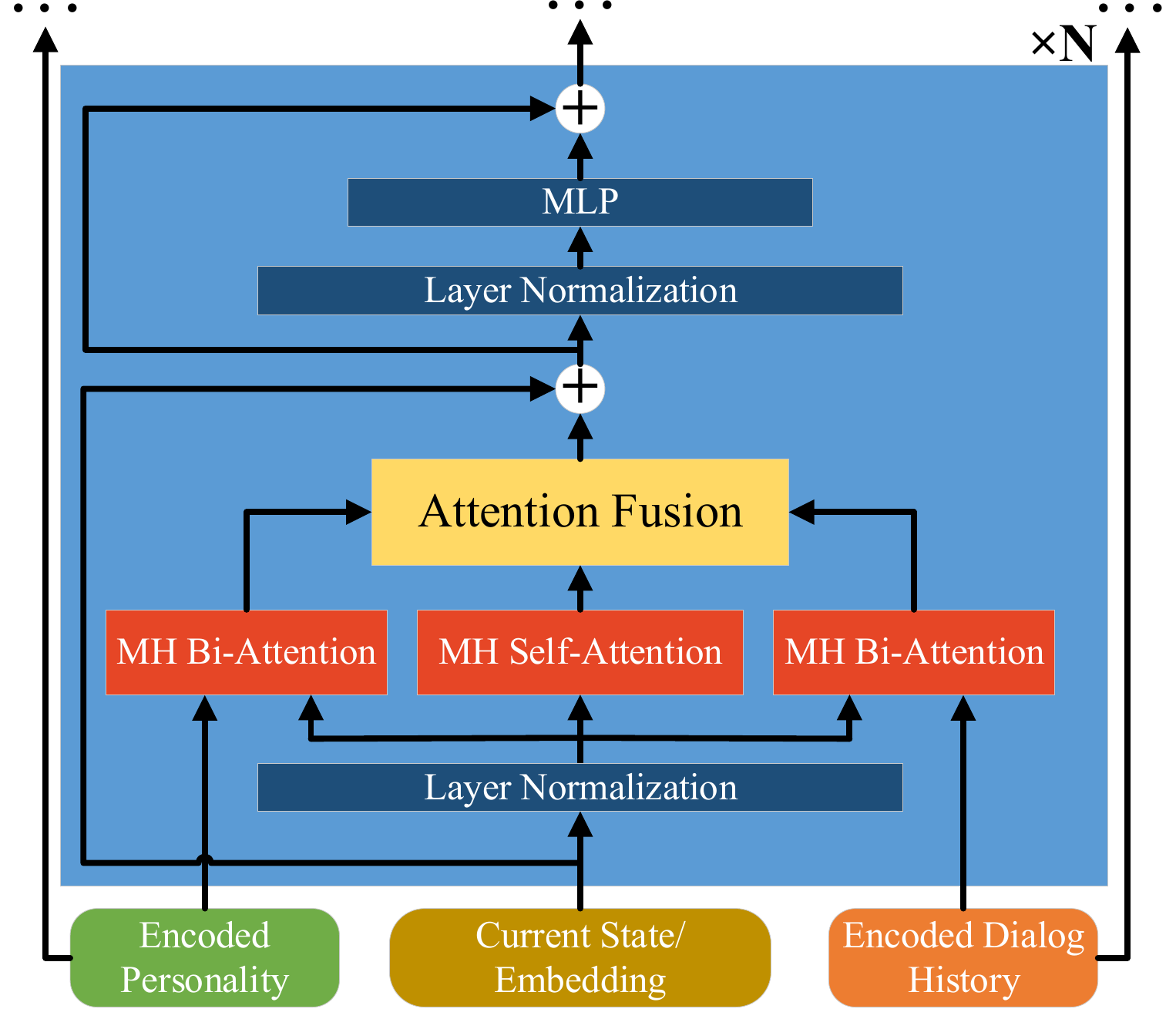}
        \label{fig:block}
    }
    \caption{Architecture of our proposed model.}
\end{figure}

Following the former work~\cite{gpt-persona}, we use the personalized dialogue generation task on PersonaChat~\cite{personachat} as an example in our study. The pretrained language model GPT2 and its parameters are duplicated to form an encoder-decoder architecture shown in Figure~\ref{fig:framework}. We use GPT2 here due to its large-scale pre-training corpus than other models and strong performance in other generation tasks.

We have three separate inputs: personal profile, dialogue history, and current reply (or previously generated response during the inference stage). Embeddings of the former two, which contain embeddings of tokens, positions as well as token types, will be successively put into the encoder, which is a GPT2 model with no attention mask to fit the encoding procedure. 
%And their encoded states are obtained respectively. 
The encoded representations, together with embeddings of current response tokens will then be used as the input of a modified GPT2 decoder. 
Each decoder block will attend the current state to the three sources using different attentions, then fuse their resulting information as input for the next layer.

Inspired by multi-task learning~\cite{multitask}, we further separate the original loss in language modeling into three parts corresponding to three input sources respectively. By applying the same linear prediction layer on the output of both encoder and decoder, three cross-entropy losses between predicted logits and corresponding truth sequences will be weighted by hyperparameters.
\begin{equation}
    \mathcal{L} = \alpha\mathcal{L}_{persona} + \beta\mathcal{L}_{history} + \gamma\mathcal{L}_{pred}
\end{equation}

with Adam optimizer~\cite{adam}.

\subsection{Block Details in Decoder}
% \footnote{***this part needs to be shorten, as i guess this part is the same as in~\cite{gpt-persona}.}
Recall that we have three input sources in the decoder, and thus some modifications are needed if the decoder structure is inherited from GPT2.
%so as to accept multiple inputs at the same time~\cite{transformer_average}.
Details of each modified decoder block are shown in Figure~\ref{fig:block}, in which the most apparent change is the additional two multi-head (MH) bidirectional attentions and the attention fusion module that fuses various attention outputs. 
% \footnote{***multi-source attention?}
The other parts remain the same as GPT2. In the following, we will first describe the MH Bi-attention. Attention fusion will be discussed in the next section. 

The MH self-attention in Transformer~\cite{transformer} handles a single input only. In order to make it accept two input sources, we regard the current state ${\bf H}^c \in \mathbb{R}^{L^c \times d}$ from the previous layer (or embedding of reply in the first layer) as query and encoded state of auxiliary information ${\bf H}^a \in \mathbb{R}^{L^a \times d}$ as key and value in the attention. Here $L^c$ and $L^a$ are corresponding lengths for these input, and ${\bf H}^a$ can be encoded personality ${\bf H}^p$ or dialog history ${\bf H}^h$. The output of each single head in MH Bi-attention can be obtained via
\begin{equation}
    {\bf A} = \mbox{softmax}(\frac{({{\bf H}^c}{{\bf W}^Q})({{\bf H}^a}{{\bf W}^K})^{\mathrm T}}{{\sqrt d }})({{\bf H}^a}{{\bf W}^V}),
\end{equation}
% \begin{eqnarray*}
%     {\bf A} = \mbox{softmax}(\frac{({{\bf H}^c}{{\bf W}^Q})({{\bf H}^e}{{\bf W}^K})^{\mathrm T}}{{\sqrt d }})({{\bf H}^e}{{\bf W}^V}),
% \end{eqnarray*}
where ${\bf W}^Q$, ${\bf W}^K$, ${\bf W}^V$ are learnable matrices. In our model, different attentions own separate parameters instead of sharing.
This differs from the previous work~\cite{gpt-persona} which reuses the self-attention for bi-attention.
Besides, the original GPT2 is a single-directional model using a triangular matrix as the attention mask. Since the auxiliary information ${\bf H}^a$ is visible for the current reply at all time steps, no mask exists in MH bi-attention. 

In total, three attention information ${\bf A}^c$, ${\bf A}^p$, and ${\bf A}^h$ are obtained by attending the current state to itself, personality, and history respectively, all in the same dimension $\mathbb{R}^{L^c \times d}$. They need to be fused into one matrix ${\bf A}^f \in \mathbb{R}^{L^c \times d}$ so as to proceed to subsequent decoding layers.

\subsection{Attention Fusion}
\label{sec:attention_fusion}
In this section, we discuss various methods to fuse the multiple attention information obtained above.
The simplest approach is to average three sources in all dimensions~\cite{gpt-persona}, which treats all sources equally. However, in different dialogues, we may need to concentrate more on the dialogue history or the persona profile in order to generate proper responses. Here we introduce the following three kinds of methods to allow for more flexible information fusion from all input sources.
\begin{itemize}[wide=0\parindent,noitemsep,topsep=0em]
\item {\bf Static methods} fuse different information using an identical fusion function with no training parameter. Except the average pooling {\bf(avg)} which is regarded as a very simple fusion baseline, we also include Maximum {\bf(max)}, and  Minimum {\bf (min)} operation for every dimension among all sources. 
 
\item {\bf Weighting methods} 
try to estimate the global optimal proportion of each source in a given domain by introducing extra learnable weights which are then fixed in inference. Such methods can be: 

\noindent (i) source-level scalar weights {\bf (sw)}, which means there are three trainable scalars $w^c, w^p, w^h$ for each source in each layer and ${\bf A}^f=(w^c{\bf A}^c+w^p{\bf A}^p+w^h{\bf A}^h)/(w^c+w^p+w^h)$. 

\noindent (ii) source-dimension level {\bf (dw)}, in which weights are learnable vectors ${\bf w}^c, {\bf w}^p, {\bf w}^h \in \mathbb{R}^d$. For each row $j$ of ${\bf A}^f$ and weight vectors $\bf w$, we perform the weighted combination via ${\bf A}_j^f=(w_j^c{\bf A}_j^c+w_j^p{\bf A}_j^p+w_j^h{\bf A}_j^h)/(w_j^c+w_j^p+w_j^h)$.

\noindent (iii) linear method ({\bf linear}) in which a linear network is used to transform the concatenated attention $[{\bf A}^c;{\bf A}^p;{\bf A}^h]$ into ${\bf A}^f$. Different from above one, each dimension in the new feature space here  contains information from all dimensions of all sources to realize a better interaction.
\item {\bf Attention-based method} fuses the information based on a trainable modified transformer attention ({\bf att}). The attention fusion function changes according to multiple input information as follows 
\begin{equation}
    {\bf Z} = \mbox{softmax}(\frac{\mbox{sign}({{{\bf A}^c}{{\bf A}^p}^{\mathrm T}}) \odot ({\sqrt {|{{\bf A}^c}{{\bf A}^p}^{\mathrm T}|}}}{{\sqrt d }}){\bf A}^h, 
\end{equation}
where $\mbox{sign}(\cdot)$ is a function with value 1 when the element is positive or -1 when negative; $|\cdot|$ for absolute value; square root ensures that the value scale remains the same. 
This method utilizes matrix multiplication to make fully interaction between all state values, obtaining the states conditioned on all information sources dynamically. History information is selected as the ``value" term to get more dialog history involved in the obtained state.
\end{itemize}

\section{Experiment}

We employ the PersonaChat~\cite{personachat, convai2} dataset in our experiments which has 164,356 utterances in 10,981 dialogues and 1,155 personas.
%and there are 7801 samples for validation and 7512 samples for test. 
Each sample contains dialog history with up to 15 utterances, a gold reply and a persona description with no more than 5 sentences.

Four kinds of dialogue models using pretrained language models as the initialization are compared: 

\noindent (i) {\bf TransferTransfo}~\cite{transfertransfo}, a single-input OpenAI GPT using token type embedding to distinguish different parts of a single concatenated input (persona profile, dialog history, and reply successively). We also replace original GPT in this method with GPT2, denoted as {\bf TransferGPT2}.  

\noindent (ii) {\bf MI-GPT}~\cite{gpt-persona} which uses the OpenAI GPT in both encoder and decoder with average pooling as the attention fusion method.
%(iii) Another single-input baseline {\bf TransferGPT2} which is similar to TransferTransfo but using GPT2 as base model, 

\noindent (iii) Our architecture using GPT2 as the base model and average as fusion method ({\bf GPT2-avg}), a very simple baseline inherited from MI-GPT.
% \footnote{***what is the model difference between yours and MI-GPT, besides the loss and gpt/gpt2? \color{red}{I illustrate it in sec2.2}}

\noindent (iv) Our model with each of the attention fusion methods discussed in Sec~\ref{sec:attention_fusion}, denoted as {\bf GPT2-X}, and {\bf X} is the corresponding fusion method.

All GPT2 models used here are small size (12 layers, hidden size is 768).
Besides, Seq2seq model with attention~\cite{seq2seq} using 6-layer Transformer as the encoder and decoder is also included as an end-to-end single-input baseline.\footnote{Source code is available on: \url{https://github.com/caoyu-noob/Multi-GPT2}}

The following automatic metrics are considered in our evaluation: BLEU~\cite{bleu}, METEOR~\cite{meteor}, NIST-4, which indicate the gram-level similarity between the references and generated responses. Moreover, Entropy-4, corpus-level Distinct-2 and the average length of replies are used to reflect the diversity of obtained text. In addition, human evaluation is also conducted on 200 dialog pairs in terms of fluency (range: $1\sim3$), relevance with dialogue history (h-rel, range: $1\sim3$) and consistency with personality (p-consist, \{0, 1\}). 
% \footnote{***how about using the term "relevance" instead of "correlation"? "correlation" may raise confusion towards the statistical correlation. }
More experiment configurations can be found in Appendix~\ref{appendix:experiment}.

\begin{table*}[t!]
\setlength{\tabcolsep}{1.3mm}
% \setlength{\abovecaptionskip}{0.2cm}
% \setlength{\belowcaptionskip}{-0.3cm}
% \small
\centering
\begin{tabular}{l|ccc|ccc|ccc}
\hline 
\textbf{Model} & BLEU & METEOR & NIST-4 & Entropy-4 & Dist-2 & Avg.len & fluency & h-rel & p-consist \\
\hline
Human & - & - & - & 10.725 & 36.688 & 11.507 & 2.901 & 2.645 & 0.598 \\
\hline
Seq2seq & 1.769 & 6.926 & 1.028 & 6.789 & 6.356 & 8.710 & - & - & - \\
TransferTransfo & 2.054 & 7.672 & 1.183 & 8.429 & 17.917 & 7.824 & 2.748 & 2.348 & 0.542 \\
MI-GPT & 3.151 & 8.112 & 1.264 & 8.054 & 13.264 & 9.026 & 2.809 & 2.150 & 0.628 \\
TransferGPT2 & 2.273 & 7.872 & 1.194 & 8.263 & 16.444 & 8.036 & 2.785 & \bf 2.385 & 0.548 \\
GPT2-avg & 3.211 & 8.149 & 1.291 & 7.904 & 13.612 & 8.932 & \bf 2.838 & 2.149 & \bf 0.648 \\
\hline
GPT2-max & 3.344 & 8.156 & 1.206 & 8.175 & 14.104 & 8.750 & - & - & - \\
% GPT2-min & 3.774 & 8.661 & 1.388 & 8.099 & 14.925 & 9.209 & \bf 2.817 & 2.213 & \bf 0.603 \\
GPT2-min & 3.774 & 8.661 & 1.388 & 8.099 & 14.925 & 9.209 & - & - & - \\
GPT2-sw & \bf 3.949 & \bf 8.881 & \bf 1.407 & 8.228 & 15.294 & 9.068 & \bf 2.814 & \bf 2.355 & 0.595 \\
GPT2-dw & 3.714 & 8.694 & 1.385 & 8.096 & 14.647 & 9.095 & - & - & - \\
GPT2-linear & \bf 4.147 & \bf 8.988 & \bf 1.408 & 8.279 & 15.237 & 9.011 & 2.777 & 2.332 & \bf 0.602 \\
GPT2-att & 3.659 & 8.449 & 1.249 & 8.028 & 14.592 & 8.723 & - & - & - \\
% GPT2-att & 3.659 & 8.449 & 1.249 & 8.028 & 14.592 & 8.723 & 2.737 & \bf 2.383 & 0.487 \\
\hline
\end{tabular}
\caption{\label{tab:results}Dialogue generation performance comparison of different models on the test set of PersonaChat. Values for BELU, METEOR and Dist-2 are in percentage. Human evaluation is only conducted on representative models.}
\end{table*}

\subsection{Results}
Results of different models on both automatic metrics and human evaluations are shown in Table~\ref{tab:results}.

%\footnote{***not all columns have bold numbers? \color{red}{The reference also didn't have bold on diversity related metrics}}
%
%, including 4 baselines and 6 proposed attention fusion methods. 
We first analyze results on automatic metrics. 
It can be observed that GPT2 is more powerful than OpenAI GPT under the same architecture. 
Multi-input (MI) models that use the encoder-decoder frameworks generally outperform single-input (SI) models (TransferTransfo, TransferGPT2) which simply concatenate all inputs. Although SI models show higher diversity, their generated texts are generally shorter. 
All attention fusion methods of our model make improvements compared to its baseline GPT2-avg. 
Among them, weighting methods have higher scores than the other two kinds of fusion methods on most metrics.
%The linear and source weighting (sw) methods deliver the best performance as learnable parameters ensure the best proportion of each source in state after fusion. On the other hand, 
Compared with static methods, weighting methods are more flexible to combine proper proportions of each source, thus it is no surprise that they can outperform static methods.
%fuse various sources roughly, while 
Meanwhile, though the attention-based method also allows for non-static attention fusion, it essentially poses dynamic weights on the history state, and thus information of persona and reply is not directly used in the final fused representation and results in its failure
It is also interesting to find that GTP2-dw shows no improvement compared to GPT2-sw, despite it extends the latter one using different weights for each dimension.
%Besides, we tried to extend sw into dimensional weights but no promotion appears.
%\footnote{***not a solid explanation. rethink. \color{red}rewritten}
%which are 0.9 and 0.7 higher in BLEU than GPT2-avg. 
%, while other fusion variants improve the auto metrics in smaller ranges. 
%Furthermore, attention fusions also raises the diversity of generated text, despite lower than SI models.

% \footnote{***why not all human results available? need to explain.}
Now we discuss human evaluation results. Here, we only conduct human evaluations on baselines and proposed models with the best automatic evaluation results (i.e. weighting methods). Fluency scores of generated texts are very close to each other even compared to gold replies, which should be largely benefited from the pretrained model. 
However,  h-rel scores (the relevance between dialog history and current responses) by models are significantly lower than those by a human. 
Note that compared with SI models, MI models using the average fusion (MI-GPT, GPT2-avg) show lower h-rel scores, though their persona consistency increases much.
This is also discussed in~\citeauthor{gpt-persona}~\citeyear{gpt-persona},
%that SI models generally have higher history correlation than MI baselines, while MI models provide more personality information in predictions with higher p-consist score. 
and the reason is that SI model is similar to a language model which stays tightly with history, while MI models take persona as a separate input which is easier to reuse personalized word. 
%Thus SI models show higher h-rel scores than MI models using avg to combine information.
%.
% However, \footnote{***should state first that MI with avg decreases the h-rel compared with SI} 
However, our models with the weighting fusion methods can not only improve the persona consistency compared to SI models, but also maintain comparable best history relevance. 
The case study of generated replies is given in Appendix~\ref{appendix:case_study}.
%comparable history relevance to SI models, at the cost of very slight drop on fluency and persona consistency.

% \footnote{***what is the connection with prev para? cannot figure out.}Treating information sources equally is not reasonable since they have different importance under diverse conditions or even each model layer. Therefore, our proposed attention fusion methods shows obvious improvement on history correlation due to learning the optimal combination of different kinds of information. And there are only very slight losses on fluency and persona consistency.

\subsection{Influence of Attention Fusion}
In this section, we further investigate how attention fusion affects the generation results, especially why using the average fusion decreases the performance on the relevance between dialog history and generated responses while the weighting fusion methods can survive. 
%So how does the attention fusion affect the generation so as to promote the history relevance? 
%We assume that the appropriate proportions of each information source in the fused results benefit the predictions. 

We group the 200 testing samples for human evaluation by their lengths of history, and then compare the average results on {\bf h-rel} scores of different methods within each group. Results are shown in Table~\ref{tab:case}.
We first compare the weighting fusion methods with the average fusion baseline.
As can be seen, all methods perform comparably when dialogue history is short.
With longer dialog history, models with weighting fusion methods perform much better than GPT2-avg. 
The reason is that when dialogue history gets longer, the effect by each history token on current reply in bi-attention is averaged out by dialogue history length, making the average fusion method harder to capture key information from any history token to generate the response. 
%Results in Table~\ref{tab:case} confirms this point. 
%We can see that all fusion methods perform comparably when history is short.
%With long dialog history, models with refined fusion perform much better than GPT2-avg. 
Next, we compare the GPT2 with weighting fusion methods to TransferGPT2 (the SI model with GPT2) and results indicate that they can also outperform SI models when dialogue history is long.
%It can also be found that fusion models have similar performance to SI models, while 
Finally, we can see that SI models beat the MI baselines with the average fusion under all conditions, proving the ineffectiveness of the simple average between different information sources.

\begin{figure}[!t]
    \centering
    \includegraphics[width=1\columnwidth]{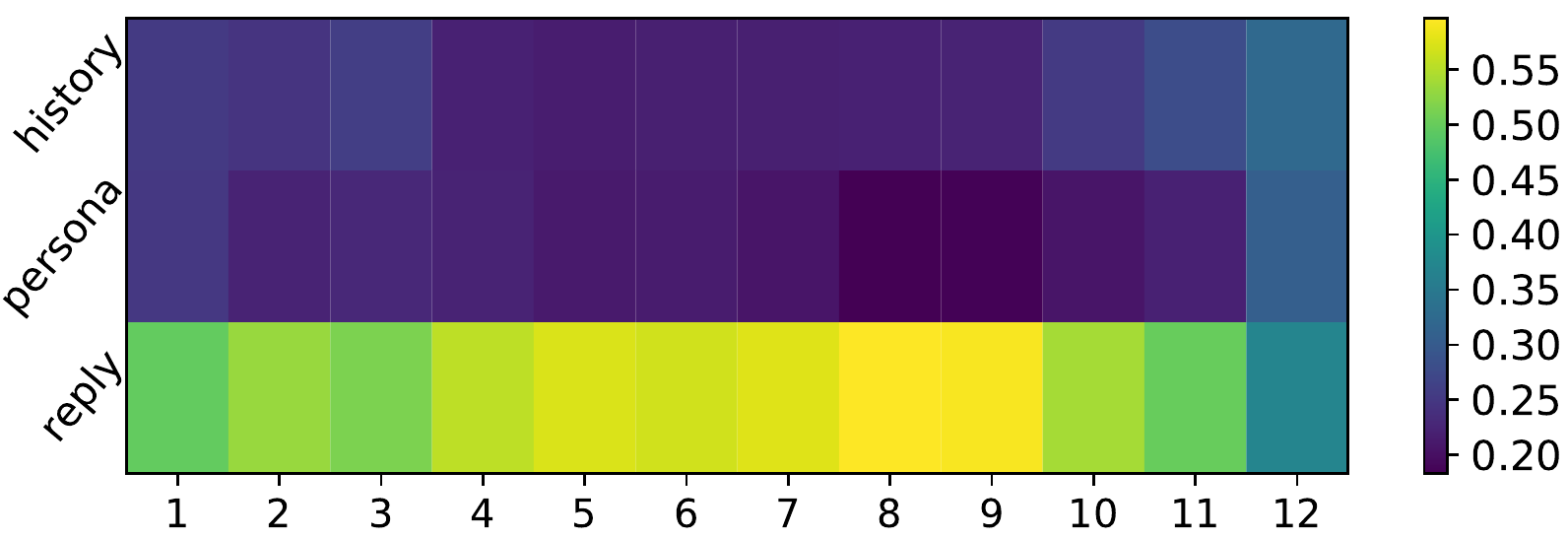}
    \caption{\label{fig:att_weight}Visualization of normalized scalar attention weights on 3 different input sources for each layer in GPT2-sw decoder.}
\end{figure}

%which is consistent with results in Table~\ref{tab:results}.
Figure~\ref{fig:att_weight} further illustrates the estimated optimal weights of each attention information in every decoder layer in GPT2-sw. We observe that attention weights of different input sources are not equal and change over different decoder layers, validating that the use of average fusion is over-simplified. The weights of diverse sources tend to be equivalent in high layers while they differ significantly in lower layers because the history and persona information are already encoded and highly abstractive.
% On one hand, accepting separate personality input makes it possible to directly reuse information from this source. We assume that the appropriate proportions of each information source in the fused results benefit the predictions. Average of multiple attentions may pose too high significance on personal profile which makes the generated text tends to demonstrate personality instead of replying the last utterance. 

%  When history is short they performs equivalently. Besides, it shows that fusion models are also better than SI models with long history, but worse when it is short due to intrinsic property of models. And SI models beat raw MI models under all conditions. More detailed case study can be found in Appendix. Moreover, the weights for current input and history is larger than for personality as shown in Figure~\ref{fig:att_weight} except the last layer in which 3 weights are similar. It illustrates that refined fusion models focus more on generate a contextual-appropriate reply.
\begin{table}[t!]
\setlength{\tabcolsep}{1.5mm}
    % \small
    \centering
    \begin{tabular}{c|c|ccc}
        \hline
         & {\small History} & \bf Win & \bf Tie & \bf Lose  \\
        \hline
        \multirow{3}*{\tabincell{c}{GPT2-weight \\ {\bf vs.} \\ GPT2-avg}}& L & 53.2\% & 28.2\% & 18.6\% \\
        ~ & M & 37.0\% & 31.1\% & 31.9\% \\
        ~ & S & 29.3\% & 45.2\% & 25.5\% \\
        \hline
        \multirow{3}*{\tabincell{c}{GPT2-weight \\ {\bf vs.} \\ TransferGPT2}} & L & 39.7\% & 35.5\% & 24.8\% \\
        ~ & M & 28.9\% & 37.1\% & 34.0\% \\
        ~ & S & 24.1\% & 43.7\% & 32.2\% \\
        \hline
        \multirow{3}*{\tabincell{c}{MI baselines \\ {\bf vs.} \\ SI baselines}} & L & 17.7\% & 30.1\% & 52.2\% \\
        ~ & M & 22.2\% & 28.9\% & 48.9\% \\
        ~ & S & 18.9\% & 42.8\% & 38.3\% \\
        \hline
    \end{tabular}
    \caption{Percentage of generated replies by the upper model better, equal or worse than the bottom one on {\bf h-rel} metric. Samples are grouped by dialog history length (long (L) / short (S) / medium (M) history length: $>$ 9 utterances / $\leq$ 3 utterances / rest samples.). GPT2-weight: GPT2-sw and GPT2-linear, MI baselines: GPT-MI and GPT2-avg, SI baselines: TransferTransfo and TransferGPT2.}
    \label{tab:case}
\end{table}

\section{Conclusion}
% To handle dialogue generation with multiple input sources, we adapt the pretrained language model GPT2 to an encoder-decoder architecture with multiple attentions from different input sources. 
% %It takes extra information as an independent input to easily reuse these profiles in predictions while it pose better importance on context history than simple average approach. 
% We investigate several attention fusion methods to obtain a better representation for generation. Experiments illustrate that weighting methods improve both automatic metrics and human-annotated dialog history relevance scores compared with baselines using simple concatenation or average fusion.
%It takes extra information as an independent input to easily reuse these profiles in predictions while it pose better importance on context history than simple average approach. 

To handle dialogue generation with multiple input sources, we adapt the pretrained language model GPT2 to an encoder-decoder architecture with multiple independent attentions for different input sources in the decoder. 
We then investigate several attention fusion methods to obtain a preferable representation for dialogue generation. Experiments illustrate that weighting methods promote both auto metrics and dialog history relevance scores annotated by human than baselines using average fusion, while they still maintain the persona consistency scores which outperform single-input models. And such architecture can be extended to other multi-input dialogue generation tasks having different information source number.

\section{Acknowledgement}
This work was supported by Australian Research Council Projects under grants FL-170100117, DP-180103424, and IC-190100031.

\bibliographystyle{acl_natbib}
\bibliography{anthology,emnlp2020}

\appendix

\section{Experiment Details}
\label{appendix:experiment}

We use the official code for the implementation of TransferTransfo~\cite{transfertransfo} and GPT2-MI~\cite{gpt-persona}, following all default settings to fine-tune models. To implement our TransferGPT2, GPT2-avg, and all refined attention fusion model, we utilize HuggingFace Transformers library\footnote{https://github.com/huggingface/transformers} with the small-size GPT2 model which has 12 layers and 768 dimensions in the hidden state. It is noted that although both our encoder and decoder are initialized from GPT2 model, their parameters are not shared. Similarly, 3 different attention modules in each layer of the decoder (1 self-attention, 2 bi-attention) are also initialized by the attention module of the corresponding layer in original GPT2 model but parameters are also not shared among them. The parameters of the additional attention fusion module will be initialized by: 1) uniform initialization for source-weighting methods, and 2) random initialization with normal distribution for linear and attention-based methods. And the linear prediction layer has the shared weight with the embedding layer of the decoder.

During fine-tuning, we use Adam optimizer~\cite{adam} with an initial learning rate 5e-4 with 0.002 warmup proportion and then a linear decay. The learning rate for the additional attention fusion module is 5$\times$ current learning rate for other parts. We train it for 5 epochs using mini-batch with size 256. And only the latest 7 utterances in dialog history are remained to avoid exceeding maximum input length. All hyperparameters are determined by manually tuning according to auto metrics BLEU, METEOR ,and NIST as criteria.

During inference, we use beam search with size 3 for all test models. Length penalty~\cite{length_penalty} is added to ensure the diversity of generation. A single NVIDIA V100 GPU with CUDA10 is used to run experiments.

\section{Case Study}
\label{appendix:case_study}

We list dialogue generation results of TransferGPT2, GPT2-avg, GPT2-sw and GPT2-linear under some cases from PersonaChat dataset~\cite{personachat} in Table~\ref{tab:case1} and Table~\ref{tab:case2}, containing samples with varied dialog history lengths. h-rel and p-consist indicate the human evaluation scores for dialogue history relevance and personality consistency of generated replies respectively. 

It can be found that our refined attention fusion models generally show similar personality consistency with the baseline GPT2-avg model who uses the same architecture but a simple average method to combine different information sources. When dialog history is long, TransferGPT2 tends to directly respond to the last history utterance using some general replies, while GPT2-avg tends to directly copy personal information as replies. GPT2-sw and GPT2-linear can properly make a response to the last context as well as involve personal profile. 
In addition, we find that when history length is not so long (length is 5 or 7), such difference will be reduced. But when dialog history is very short (less than 5), all encoder-decoder models tend to generate universal replies or simply reuse personalities because the history information is too limited for them to combine it with the given personal profile. While the single-input TransferGPT2 is inclined to reuse personality descriptions because the whole input sequence length is shorter and persona information obtains more attention compared to the input having a long history.

\begin{table*}[]
    \centering
    \begin{tabular}{|c|c|c|c|}
        \hline
        {\bf item} & {\bf text} & {\bf \color{red} h-rel} & {\bf \color{blue} p-consist} \\
        \hline 
         Personality & \tabincell{c}{ i have one cat. \\ i am a kindergarten teacher. \\ i can barely pay my bills every month.\\ i share an apartment in seattle with two roommates. \\ i just graduated from college two years ago.} & & \\
        \hline
        \tabincell{c}{Dialog history \\ (length=13)} & \tabincell{l}{
            ... \\
            A: i also love to travel and take adventures. what are\\ your hobbies? \\
            B: i like to hang out with friends \\
            A: nice. where do you like to go? \\
            B: we like to do different things do you have pets? \\
            A: i have a dog. his name is max} & & \\ 
         \hline
         Human & {\color{red} that is cute} i have {\color{blue} a cat}. & 3 & 1 \\
         \hline
         TransferGPT2 & that is a {\color{red} cute name}. & 3 & 0 \\
         \hline
         GPT2-avg & i have two {\color{blue} cats} and a dog. & 2 & 0.5 \\
         \hline
         GPT2-sw & {\color{red} that is cool} i have two {\color{blue} cats}. & 3 & 0.5 \\
         \hline
         GPT2-linear & that is a {\color{red} cute name} i have {\color{blue} a cat named maxy}. & 3 & 1 \\
         \hline
         \hline
         Personality & \tabincell{c}{my favorite tv show is the office. \\
            i do not drive because i live in new york. \\
            i like jazz music. \\
            i do not eat anything sweet.} & & \\
         \hline
         \tabincell{c}{Dialog history \\ (length=9)} & \tabincell{l}{
            ... \\
            A: i need something to mellow me out after working all day \\
            B: that can be hard to find up in new york.\\things are fast and tight here \\
            A: do you have many insects? i do not care for that. \\
            B: not in the city, but yes upstate. \\
            A: what do you do for a living? i am in the science\\ field engineer} & & \\ 
         \hline
         Human & {\color{red} i am a nutritionist} , so {\color{blue} no sweets} for me. & 3 & 1 \\
         \hline
         TransferGPT2 & {\color{red} i am a stay at home mom}. & 3 & 0 \\
         \hline
         GPT2-avg & i live in {\color{blue} new york city.} & 2 & 1 \\
         \hline
         GPT2-sw & {\color{red} i do not work}. i just sit around and {\color{blue}listen to jazz music}. & 3 & 1 \\
         \hline
         GPT2-linear & {\color{red}that is cool. i work in a lab} in {\color{blue}new york city}. & 3 & 1 \\
         \hline
    \end{tabular}
    \caption{Some cases of generated dialogue replies by TrnasferGPT2, GPT2-avg, GPT2-sw and GPT2-linear.}
    \label{tab:case1}
\end{table*}

\begin{table*}[]
    \centering
    \begin{tabular}{|c|c|c|c|}
        \hline
        {\bf item} & {\bf text} & {\bf \color{red} h-rel} & {\bf \color{blue} p-consist} \\
        \hline 
         Personality & \tabincell{c}{   i am a huge fantasy fan.\\
            i want to write fantasy books myself.\\
            i have two part time jobs.\\
            i like to binge watch the lord of the rings} & & \\
        \hline
        \tabincell{c}{Dialog history \\ (length=7)} & \tabincell{l}{
            ...\\
            A: ah gosh. that is the worst! \\
            B: i want to just stay home and write. \\
            A: i would rather play softball than go to work \\
            B: i like to role play and play fantasy games. have you been\\ to com icon? \\
            A: what is that? no i have never been} & & \\ 
         \hline
         Human & {have you ever heard of {\color{blue} the lord of the rings}?}. & 2 & 1 \\
         \hline
         TransferGPT2 & that is cool. i am a {\color{blue} huge fantasy fan}. & 2 & 1 \\
         \hline
         GPT2-avg & {\color{red}it is a} {\color{blue}fantasy} game. & 3 & 1 \\
         \hline
         GPT2-sw & {\color{red} it is a} {\color{blue}fantasy world} with lots of dragons. & 3 & 1 \\
         \hline
         GPT2-linear & {\color{red} i have not}. i am more of a {\color{blue} fantasy fan}. & 2 & 1 \\
         \hline
         \hline
         Personality & \tabincell{c}{i do not like exercise or physical activity.\\
            i am a homebody. \\
            my favorite food is ice cream. \\
            i have a husband who i love very much.\\
            i love watching movies and tv.} & & \\
         \hline
         \tabincell{c}{Dialog history \\ (length=5)} & \tabincell{l}{
            A: hello ! tell me about yourself? \\
            B: i really like to stay at home at watch a movie with my \\darling hubby. you? \\
            A: that sounds nice! i am a nurse who loves zoos and teddy\\ bears. \\
            B: aww do you collect teddy bears? my sweet sweet husband\\ buys me them \\
            A: i do! i bring them to my patients at the childrens hospital.} & & \\ 
         \hline
         Human & {\color{red} so nice!} do they also {\color{blue} like ice cream? i love it}! & 3 & 1 \\
         \hline
         TransferGPT2 & i do not have any kids, but i do {\color{blue} have a husband}. & 1 & 1 \\
         \hline
         GPT2-avg & {\color{red} that's so sweet}! i love to eat {\color{blue}ice cream} and ice cream. & 3 & 1 \\
         \hline
         GPT2-sw & {\color{red} that sounds like a lot of fun}. i am {\color{blue} a homebody myself}. & 3 & 1 \\
         \hline
         GPT2-linear & {\color{red} that is nice}. i am {\color{blue} a homebody myself}. & 3 & 1 \\
         \hline
         \hline
         Personality & \tabincell{c}{ i am going to auburn university next year as a freshman.\\
                i love shopping fashion and clothes.\\
                i want to be a doctor when i grow up.\\
                i am 17 years old and a senior in high school.} & & \\
         \hline
         \tabincell{c}{Dialog history \\ (length=3)} & \tabincell{l}{
            A: hi i am talking to text\\
            B: hello , interesting introduction you have there\\
            A: colors i can not see them well , my sis helps me out} & & \\ 
         \hline
         Human & {\color{red} that is unfortunate}. what is something that you enjoy doing? & 3 & 0 \\
         \hline
         TransferGPT2 & i am a junior in {\color{blue} high school}, i {\color{blue} want to be a doctor}  & 3 & 0 \\
         \hline
         GPT2-avg & {\color{red} what do you do for a living}? & 3 & 0 \\
         \hline
         GPT2-sw & {\color{red} i am sorry to hear that.} & 1 & 1 \\
         \hline
         GPT2-linear & {\color{red} what do you do for a living}? & 3 & 0 \\
         \hline
    \end{tabular}
    \caption{Some cases of generated dialogue replies by TrnasferGPT2, GPT2-avg, GPT2-sw and GPT2-linear.}
    \label{tab:case2}
\end{table*}

\end{document}